\newif\ifIcassp
\Icassptrue
\ifIcassp
    \documentclass{article}
    \usepackage{spconf}
\else 
    \documentclass[journal,12pt,onecolumn,draftclsnofoot]{IEEEtran}
\fi 
\usepackage{cite}
\usepackage{amsmath,amssymb,amsfonts,amsthm}
\usepackage{graphicx}
\usepackage{textcomp}
\usepackage{algorithm}
\usepackage[noend]{algpseudocode}
\usepackage{url}
\usepackage[usenames]{color}
\usepackage{fancyhdr}
\usepackage{listings}
\usepackage{psfrag,caption,subcaption}
\usepackage{stfloats}
\usepackage{tabularx}
\usepackage{arydshln}

\usepackage[dvipsnames]{xcolor}
\usepackage[normalem]{ulem}
\usepackage{array}
\newcolumntype{Y}{>{\centering\arraybackslash}X}

\def\cast{{
   \mathord{
      \hbox to 0em{
         \ooalign{
	   \smash{\hbox{$\ast$}}\crcr
	   \smash{\hskip-1pt\Large\hbox{$\circ$}} }
	 \hidewidth}
      \phantom{\bigcirc}
} }}

\newcommand{\rT}{^{ \raisebox{1.2pt}{$\rm \scriptstyle T$}}}

\newcommand{\bds}{\begin {itemize}}
\newcommand{\eds}{\end {itemize}}
\newcommand{\bdf}{\begin{definition}}
\newcommand{\blm}{\begin{lemma}}
\newcommand{\edf}{\end{definition}}
\newcommand{\elm}{\end{lemma}}
\newcommand{\bthm}{\begin{theorem}}
\newcommand{\ethm}{\end{theorem}}
\newcommand{\bprp}{\begin{prop}}
\newcommand{\eprp}{\end{prop}}
\newcommand{\bcl}{\begin{claim}}
\newcommand{\ecl}{\end{claim}}
\newcommand{\bcr}{\begin{coro}}
\newcommand{\ecr}{\end{coro}}
\newcommand{\bquest}{\begin{question}}
\newcommand{\equest}{\end{question}}

\newcommand{\larrow}{{\larrow}}

\newcommand{\argmin}{\ensuremath{\mathrm{arg}\min}}
\newcommand{\argmax}{\ensuremath{\mathrm{arg}\max}}

\newcommand{\cE}{{\ensuremath{\mathcal{E}}}}

\newcommand{\cG}{{\ensuremath{\mathcal{G}}}}

\newcommand{\cN}{{\ensuremath{\mathcal{N}}}}

\newcommand{\cV}{{\ensuremath{\mathcal{V}}}}

\newcommand{\cbX}{{\ensuremath{ \boldsymbol{\mathcal{X}}}}}

\newcommand{\vc}{{\ensuremath{{\mathbf{c}}}}}

\newcommand{\vx}{{\ensuremath{{\mathbf{x}}}}}

\newcommand{\mS}{{\ensuremath{\mathbf{S}}}}

\newcommand{\mX}{{\ensuremath{\mathbf{X}}}}

\usepackage{latexsym}

\def\IC{\mathbb C}
\def\IN{\mathbb N}
\def\IZ{\mathbb Z}
\def\IR{\mathbb R}

\def\shat{^{\mathchoice{}{}%
 {\,\,\smash{\hbox{\lower4pt\hbox{$\widehat{\null}$}}}}%
 {\,\smash{\hbox{\lower3pt\hbox{$\hat{\null}$}}}}}}

\def\bSigma{{
      \ooalign{
      \smash{\hskip.4pt\raise.4pt\hbox{$\Sigma$}}\vphantom{}\crcr
      \smash{\hskip.7pt\raise.6pt\hbox{$\Sigma$}}\vphantom{}\crcr
      \smash{\hbox{$\Sigma$}}\vphantom{$\Sigma$}}
      \vphantom{\hbox{$\Sigma$}}
      }}
\def\bTheta{{
      \ooalign{
      \smash{\hskip.5pt\raise.5pt\hbox{$\Theta$}}\vphantom{}\crcr
      \smash{\hskip.0pt\raise.1pt\hbox{$\Theta$}}\vphantom{}\crcr
      \smash{\hbox{$\Theta$}}\vphantom{$\Theta$}}
      \vphantom{\hbox{$\Theta$}}
      }}
\def\bDelta{{
      \ooalign{
      \smash{\hskip.4pt\raise.4pt\hbox{$\Delta$}}\vphantom{}\crcr
      \smash{\hskip.7pt\raise.6pt\hbox{$\Delta$}}\vphantom{}\crcr
      \smash{\hbox{$\Delta$}}\vphantom{$\Delta$}}
      \vphantom{\hbox{$\Delta$}}
      }}
\def\bLambda{{
      \ooalign{
      \smash{\hskip.5pt\raise.5pt\hbox{$\Lambda$}}\vphantom{}\crcr
      \smash{\hskip.0pt\raise.1pt\hbox{$\Lambda$}}\vphantom{}\crcr
      \smash{\hbox{$\Lambda$}}\vphantom{$\Lambda$}}
      \vphantom{\hbox{$\Lambda$}}
      }}

\makeatletter

\def\bordermatrix#1{\begingroup \m@th
  \@tempdima 8.75\p@
  \setbox\z@\vbox{%
    \def\cr{\crcr\noalign{\kern2\p@\global\let\cr\endline}}%
    \ialign{$##$\hfil\kern2\p@\kern\@tempdima&\thinspace\hfil$##$\hfil
      &&\quad\hfil$##$\hfil\crcr
      \omit\strut\hfil\crcr\noalign{\kern-\baselineskip}%
      #1\crcr\omit\strut\cr}}%
  \setbox\tw@\vbox{\unvcopy\z@\global\setbox\@ne\lastbox}%
  \setbox\tw@\hbox{\unhbox\@ne\unskip\global\setbox\@ne\lastbox}%
  \setbox\tw@\hbox{$\kern\wd\@ne\kern-\@tempdima\left[\kern-\wd\@ne
    \global\setbox\@ne\vbox{\box\@ne\kern2\p@}%
    \vcenter{\kern-\ht\@ne\unvbox\z@\kern-\baselineskip}\,\right]$}%
  \null\;\vbox{\kern\ht\@ne\box\tw@}\endgroup}
\makeatother

\makeatletter
\def\argmin{\mathop{\operator@font arg\,min}}
\def\argmax{\mathop{\operator@font arg\,max}}
\makeatother

\newcommand{\bea}{\begin{array}}
\newcommand{\ena}{\end{array}}
\newcommand{\beq}{\begin{equation}}
\newcommand{\enq}{\end{equation}}

\newcommand{\beqa}{\begin{eqnarray}}
\newcommand{\enqa}{\end{eqnarray}}

\newcommand{\beqan}{\begin{eqnarray*}}
\newcommand{\enqan}{\end{eqnarray*}}

\newcommand{\AL}{\begin{enumerate}}
\newcommand{\ALE}{\end{enumerate}}

\def\addots{\mathinner{
    \mkern1mu\raise0pt\vbox{\kern7pt\hbox{.}}
    \mkern2mu\raise4pt\hbox{.}
    \mkern2mu\raise7pt\hbox{.}
    \mkern1mu}}

\def\sddots{\mathinner{
    \mkern.8mu\raise7pt\hbox{.}
    \mkern.8mu\raise4pt\hbox{.}
    \mkern.8mu\raise0pt\vbox{\kern7pt\hbox{.}}
    \mkern1mu}}

\def\saddots{\mathinner{
    \mkern.2mu\raise0pt\vbox{\kern7pt\hbox{.}}
    \mkern.2mu\raise4pt\hbox{.}
    \mkern.2mu\raise7pt\hbox{.}
    \mkern1mu}}

\def\sqplus{\mathbin{
	{\ooalign{\hfil\raise.3ex\hbox{\scriptsize
	+}\hfil\crcr\mathhexbox274\crcr\mathhexbox275}}
	}} 
\def\sqminus{\mathbin{
	{\ooalign{\hfil\raise.3ex\hbox{\scriptsize
	--}\hfil\crcr\mathhexbox274\crcr\mathhexbox275}}
	}}

\def\IC{{
   \mathord{
      \hbox to 0em{
	 \hskip-4pt
         \ooalign{
	   \smash{\hskip1.9pt\raise2.6pt\hbox{$\scriptscriptstyle |$}}\crcr
	   \smash{\hbox{\rm\sf C}} }
	 \hidewidth}
      \phantom{\hbox{\rm\sf C}}
} }}
\def\IN{
    {\ooalign{
   \smash{\hskip2.2pt\raise1.5pt\hbox{$\scriptscriptstyle |$}}\vphantom{}\crcr
   \hbox{\sf N}
	}}
	} 
\def\IZ{
    {\ooalign{
   \smash{\hskip1.9pt\raise0pt\hbox{$\sf Z$}}\vphantom{}\crcr
   \hbox{\sf Z}
	}}
	} 
\def\IR{
    {\ooalign{
   \smash{\hskip2.2pt\raise1.5pt\hbox{$\scriptscriptstyle |$}}\vphantom{}\crcr
   \smash{\hskip2.2pt\raise3.3pt\hbox{$\scriptscriptstyle |$}}\vphantom{}\crcr
   \hbox{\sf R}
	}}
	} 

\DeclareMathAlphabet{\mathcmb}{OT1}{cmr}{b}{n}

\def\bSigma{\ensuremath{\mathcmb{\Sigma}}}
\def\bLambda{\ensuremath{\mathcmb{\Lambda}}}

\def\bTheta{\ensuremath{\mathcmb{\Theta}}}

\newcommand{\SI}{\begin{indlist}}
\newcommand{\EI}{\end{indlist}}

\newcommand{\DL}{\begin{dashlist}}
\newcommand{\DLE}{\end{dashlist}}

\makeatletter
\def\setboxz@h{\setbox\z@\hbox}
\def\wdz@{\wd\z@}
\def\boxz@{\box\z@}
\def\underset#1#2{\binrel@{#2}%
  \binrel@@{\mathop{\kern\z@#2}\limits_{#1}}}
\def\binrel@#1{\begingroup
  \setboxz@h{\thinmuskip0mu
    \medmuskip\m@ne mu\thickmuskip\@ne mu
    \setbox\tw@\hbox{$#1\m@th$}\kern-\wd\tw@
    ${}#1{}\m@th$}%
  \edef\@tempa{\endgroup\let\noexpand\binrel@@
    \ifdim\wdz@<\z@ \mathbin
    \else\ifdim\wdz@>\z@ \mathrel
    \else \relax\fi\fi}%
  \@tempa
}
\let\binrel@@\relax%
\makeatother

\begin{document}

\title{Learning Sparse Graphs with a Core-periphery Structure
}
\ifIcassp
    \ninept
%
%
\name{Sravanthi Gurugubelli  and  Sundeep Prabhakar Chepuri 
\thanks{This work is supported in part by the Pratiksha Trust Fellowship and SERB SRG/2019/000619.}
}
\address{Indian Institute of Science, Bangalore, India
}
\else
\author{\IEEEauthorblockN{Sravanthi Gurugubelli and Sundeep Prabhakar Chepuri}\\
\IEEEauthorblockA{\textit{Indian Institute of Science, Bangalore, India}}}
\fi
\maketitle
\begin{abstract}
In this paper, we focus on learning sparse graphs with a core-periphery structure. We propose a generative model for data associated with core-periphery structured networks to model the dependence of node attributes on core scores of the nodes of a graph through a latent graph structure. Using the proposed model, we jointly infer a sparse graph and nodal core scores that induce dense (sparse) connections in core (respectively, peripheral) parts of the network. Numerical experiments on a variety of real-world data indicate that the proposed method learns a core-periphery structured graph from node attributes alone, while simultaneously learning core score assignments that agree well with existing works that estimate core scores using graph as input and ignoring commonly available node attributes.
\end{abstract}

\vspace*{-2mm}
\section{Introduction}

Mesoscale properties of graphs are often used to capture the structure of complex networks. A prevalent mesoscale feature in real-world networks is the core-periphery structure~\cite{Borgatti2000Models}. 
Core-periphery property is ubiquitous in social networks~\cite{barbera2015the}, trade and transport networks~\cite{Verma2016Emergence}, citation networks, and communication networks~\cite{Alvarez2005K}. They are useful in analyzing biological networks like brain networks~\cite{bassett2013task,Harlalka2019Atypical}, genome-scale metabolic networks of organisms\cite{Da2008Centrality}, and network of protein–protein interactions\cite{Jeong2001Lethality}, to name a few. For instance, in brain networks, the core-periphery structure explains cognitive learning processes~\cite{bassett2013task}
and in social networks~\cite{barbera2015the} (contact networks~\cite{Kitsak2010Identification}), the most influential spreaders of information (respectively, disease) are observed to be in the core part of the network.
Therefore, identifying the core and peripheral vertices helps in analyzing the central processes in complex networks.  

A core-periphery structure in graphs refers to the presence of densely connected groups of core vertices and sparsely connected periphery vertices. Core vertices are those vertices that have cohesive connections among them. Peripheral vertices, on the other hand, are not well connected to each other but are relatively well connected to core vertices. An example graph with a core-periphery structure is shown in Fig.~1(a). The dark nodes in the figure with sparse connections are peripheral nodes, while the lighter ones with dense connections are core nodes. Fig.~1(b) shows the adjacency matrix corresponding to the graph with vertices ordered in the descending order of coreness, where the lower-right block of the matrix corresponds to periphery-periphery connections and can be observed to be very sparse compared to the upper-left block corresponding to core-core connections. 

\begin{figure}
    \centering
    \includegraphics[width=0.8\columnwidth]{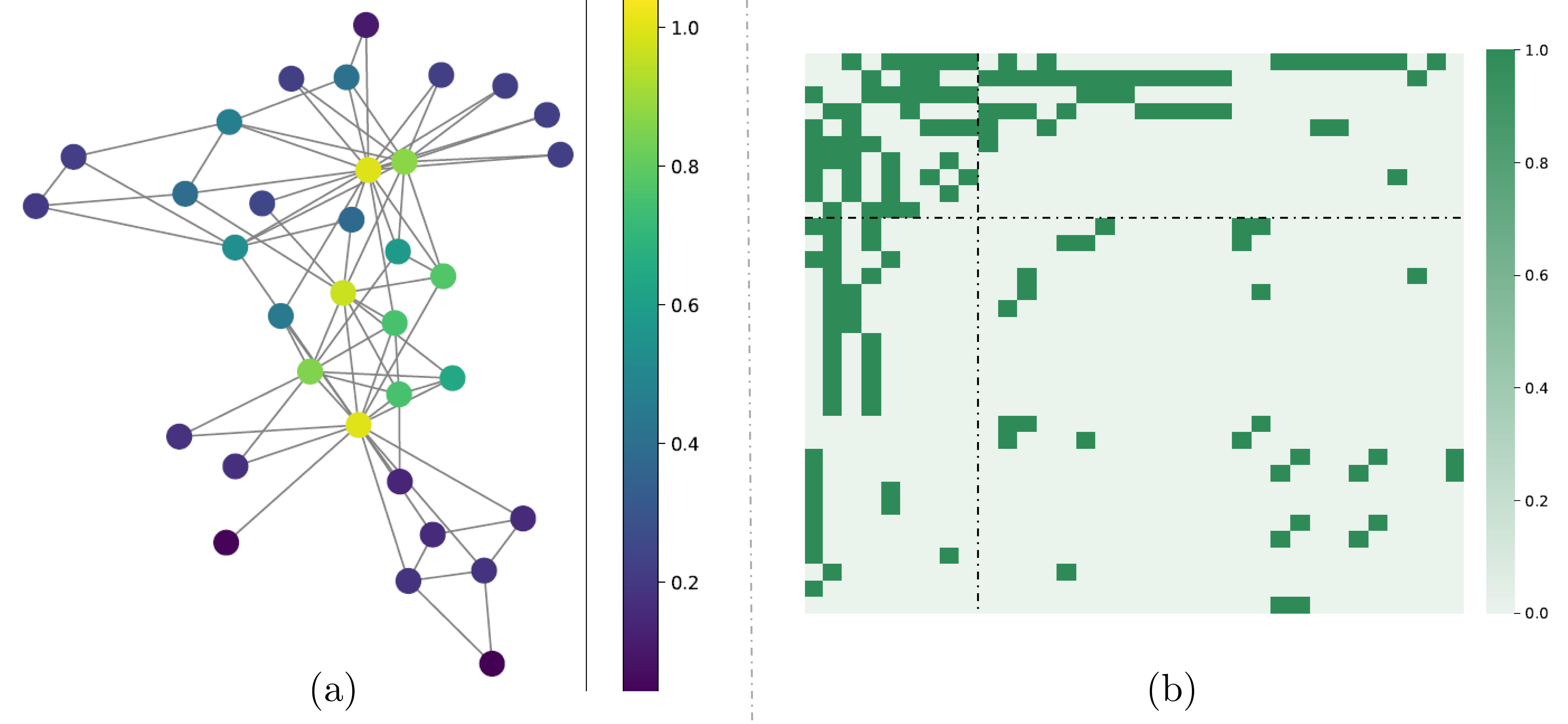}
    \caption{\small (a) A network with core-periphery structure and (b) its adjacency matrix ordered according to decreasing vertex core score.}
    \label{fig:karate eg}
\vspace*{-2mm}  
\end{figure}
\begin{figure}
    \centering
    \includegraphics[width=0.8\columnwidth]{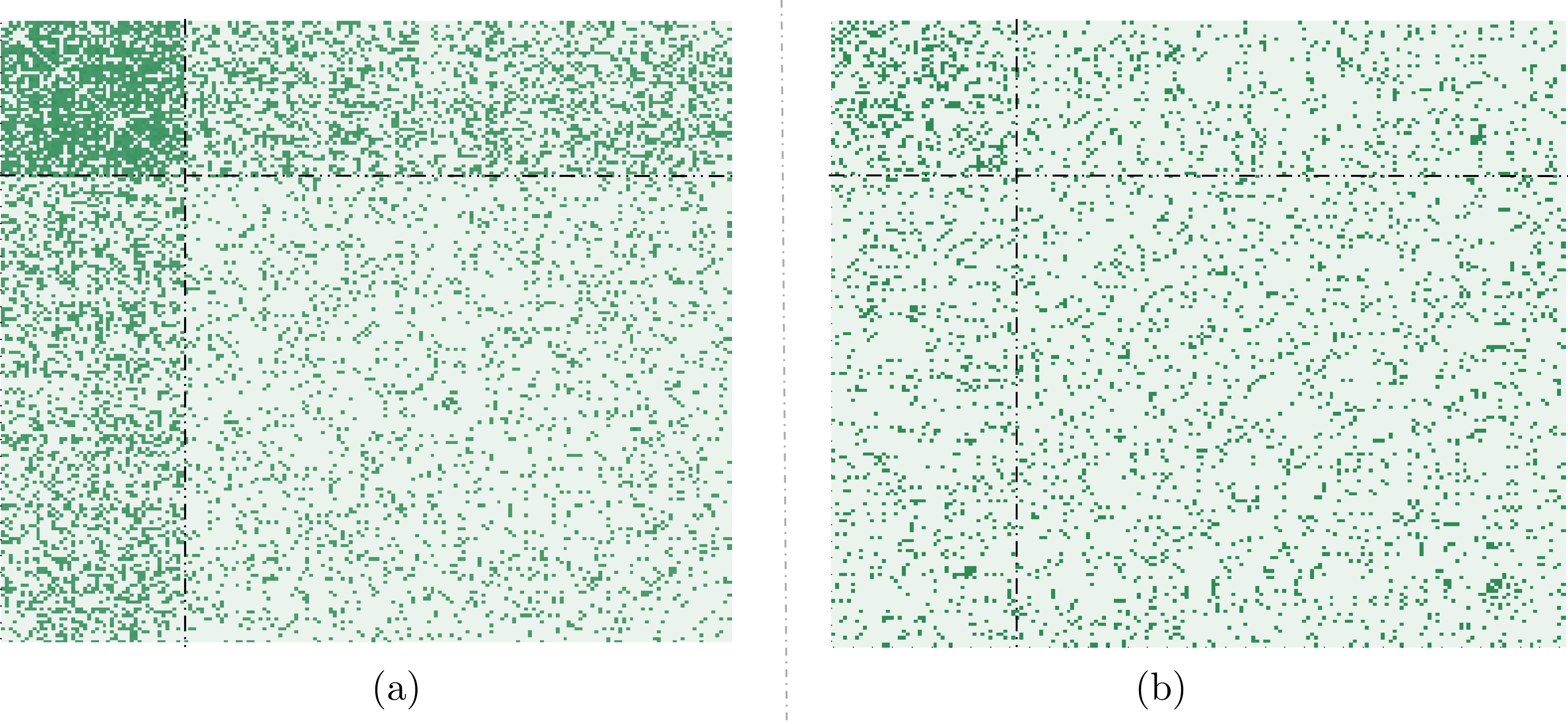}
    \caption{\small Adjacency matrices estimated from (a) the proposed method and (b) graphical lasso on a webpage network dataset~\cite{Craven1998Learning}.}
    \label{fig:webpage}
\vspace*{-4mm}  
\end{figure}
Existing algorithms estimate core scores of the nodes of a graph given the network topology~\cite{Borgatti2000Models,Rombach2014Core,Holme2005Core, Della2013Profiling,Jia2019Random,Zhang2015Identification}, but ignore node attributes that might also have information about the coreness of nodes. In many applications, we have access only to attributes of entities, and the underlying graph structure may not always be available. For example, in brain network analysis, we may have functional magnetic resonance imaging (fMRI) data of different subjects without information about the underlying structural connectivity. Therefore, in this work, we develop an approach that learns a core-periphery structured graph from node attributes alone so that the coreness of nodes are revealed implicitly.
Conventional approaches to network topology inference~\cite{dong19LearnGraphData,Sravanthi2020Learning,product2020Sai} are based on graphical lasso, which assumes a Gaussian Markov random field model for data and estimates a conditional independence graph determined by the estimated sparse inverse covariance matrix~\cite{Friedman2008Sparse}. However, the sparsity pattern recovered from graphical lasso does not readily incorporate a core-periphery structure in networks. To incorporate such a core-periphery structure, it is important to model the generative process of the edges in a graph through core scores of the vertices of the graph. See an illustration of networks estimated using the proposed method (described later on) and graphical lasso in Fig.~\ref{fig:webpage}(a) and Fig.~\ref{fig:webpage}(b), respectively. It can be observed that graphical lasso does not capture the core-periphery structure.

In this work, we propose a generative data model to relate the node attributes to the graph as well as to the nodal core scores. The proposed probabilistic generative model for node attributes models the dependence of the node attributes on the core scores through a latent graph structure. In particular, we model core scores as variables that influence the sparsity of the graph. Though often ignored in graph analysis tasks, spatial distances between nodes play a vital role in differentiating the core nodes from the peripheral ones~\cite{Jia2019Random}. For example, countries far apart in a world trade network are less likely to be connected. To this end, we also incorporate spatial information into our model. Using the proposed position-aware probabilistic model, which promotes a core-periphery network structure, we jointly estimate a sparse graph and core scores of every node in the graph. Specifically, the proposed estimator jointly learns a sparse graph structure and node core score assignments that induce dense (sparse) connections in core (respectively, peripheral) parts of the network while accounting for the spatial distances between the nodes whenever available.  We evaluate the proposed method through a number of numerical experiments on real-world data from various domains like brain, social, and transportation networks. We verify the correctness of the core scores learnt using the proposed method by comparing them with existing core score estimation algorithms, which use only the underlying known graph. The results indicate that the proposed method estimates core scores of the vertices from node attributes alone and are on par with existing methods. We also apply our method on fMRI data and report interesting observations about the differences in interactions between the brain regions in healthy subjects and individuals with attention deficit hyperactivity disorder (ADHD).

\vspace*{-2mm}
\section{Background: Gaussian graphical model}

Consider a weighted and undirected graph $\cG= \{\cV,\cE\}$, where $\cV = \{v_1, \cdots v_N\}$ is the vertex set with $N$ vertices and $\cE$ is the edge set. Let us collect the vertex attributes in the feature matrix 
$\mX = [ \vx_1, \vx_2, \cdots, \vx_d ] \in \mathbb{R}^{N\times d}$, where the $i$th row of $\mX$ contains $d$ features of the entity associated to the $i$th vertex of $\cG$. 

In a Gaussian graphical model, $\vx_1, \vx_2, \hdots, \vx_d$ are modeled as independent and identical observations drawn from $\cN(\boldsymbol{\mu},\boldsymbol{\Sigma})$, where $\boldsymbol{\mu} \in \mathbb{R}^N$ and $\boldsymbol{\Sigma} \in \mathbb{R}^{N \times N}$ is a positive definite matrix. The sparsity structure of the precision matrix $\bTheta = \boldsymbol{\Sigma}^{-1}$ encodes all the conditional dependencies between the $N$ variables associated with the vertices of $\cG$. Specifically, any $(i,j)$th entry of $\bTheta$ being zero implies conditional independence of variables associated with vertices $i$ and $j$, given the rest, and that there is no edge between the two vertices.
Graphical lasso learns the sparsity pattern in $\bTheta$ by solving an $\ell_1$-regularized Gaussian maximum log-likelihood problem~\cite{Friedman2008Sparse}:
\begin{equation}
\underset{\bTheta \succeq 0}{\text{maximize}\quad} {\rm log} \operatorname{det} \bTheta-\operatorname{tr}(\mathbf{S} \bTheta)-\lambda\|\bTheta\|_{1},
\label{eq:graphical lasso}
\end{equation}
where $\mS$ is the empirical covariance matrix and $\lambda>0$ is a regularization parameter that controls the sparsity in $\bTheta$.  Although graphical lasso recovers sparse graphical models, it does not readily incorporate any specific sparsity structure such as the core-periphery structure of interest as the $\ell_1$-penalty is uniformly applied on all the edges. 


\vspace*{-2mm}
\section{Model description}

In this section, we propose a prior that induces a sparsity pattern in graphs determined by the core scores of its vertices and the spatial distances between the vertices. Then using this probabilistic model, we propose an estimator for learning sparse Gaussian graphical models with a core-periphery structure.


Let $\vc = [c_1,c_2,\ldots,c_N]\rT \in \mathbb{R}^{N}$ denote a vector containing the core scores with $c_i\in[0,1]$ denoting the core strength of vertex~$i$. In other words, the likelihood of vertex~$i$ belonging to the core part of the network increases with the value of $c_i$. Also, let $d_{ij}$ denote the spatial distance between vertices $i$ and $j$. We now propose a probabilistic generative model that relates the node attributes in $\mX$ to its core scores through $\bTheta$. 

We model the node attributes based on a Gaussian graphical model. That is, the conditional probability distribution of $\mX$ given the precision matrix $\bTheta$, is given by (up to constants) 
\begin{equation}
    p(\mX|\bTheta) = \operatorname{det}\bTheta\,\,  {\rm exp}(-\operatorname{tr}(\mathbf{S} \bTheta)).
\end{equation}  

In networks with a core-periphery structure, we have sparser connections between vertices in the periphery, relatively denser connections between the vertices in the core and periphery, and very dense connections between vertices in the core [cf. Fig.~\ref{fig:karate eg}]. Further, vertices that are spatially well separated have sparser connections between them. To promote a sparsity pattern determined by the core-periphery structure in the graph, we therefore model the edges of the graph such that the value of $\Theta_{ij}$ is very small when $c_i + c_j - e \,{\rm log} (d_{ij})$ is small, i.e., if vertices $i$ and $j$ both belong to the periphery or if they are spatially far apart. Here, parameter $e$ controls the dependence of $d_{ij}$ on $\Theta_{ij}$.  The parameter $e$ is set to $0$ if the spatial information is not available or accounted for. To satisfy our above requirements, we model the generative process for each entry of $\Theta_{ij}$ using a Laplace distribution $p(\Theta_{ij};  c_i, c_j)$ with inverse diversity parameterized by the latent variables $c_i$ and $c_j$ as  
\begin{equation}
    w_{ij} = 1- c_i - c_j + e\, {\rm log}(d_{ij}).
\end{equation}
Specifically, the prior distribution of $\bTheta$ parameterized by $\vc$ is 
 \begin{align*}
p(\bTheta;\vc) = \prod_{i,j=1}^{N}p(\Theta_{ij};c_i,c_j)  
= Z \prod_{i,j=1}^{N} {\rm exp}\left(-\lambda w_{ij}|\Theta_{ij}|\right),
\label{eq:theta|c}
\end{align*}
where $Z$ is the normalization constant and $\lambda > 0$ controls the overall impact of $c_i$ and $c_j$ on $\Theta_{ij}$. For $p(\Theta_{ij};\vc)$ for, $i,j$, $1,\hdots,N$ to be a valid probability distribution, the inverse diversity parameter $w_{ij}$ should be nonnegative, i.e., $1 - c_i - c_j + e\, {\rm log}(d_{ij}) > 0$
for $i,j$, $1,\hdots,N$. Next, we aim to infer the model parameters $\bTheta$ and $\vc$ based on the observed node attributes~$\mX$.

\vspace*{-2mm}
\section{The proposed learning algorithm} \label{sec:learning_algo}

In this section, we present an algorithm to jointly learn the core score vector $\vc$ and a sparse graph represented by the zero pattern in~$\bTheta$. We estimate the model parameters $\bTheta$ and $\vc$ by maximizing the posterior distribution given data, i.e., by maximizing 
\begin{equation}
l(\bTheta,\vc) =  {\rm log} \, p(\mX | \bTheta) +{\rm log}\, p(\bTheta;\vc)
 \label{eq:log post. prob.} 
\end{equation} 
with respect to the parameters $\bTheta$ and $\vc$. The log-likelihood function is ${\rm log} \, p(\mX | \bTheta) = \log \operatorname{det} \bTheta-\operatorname{tr}(\mathbf{S} \bTheta)$. The prior distribution ${\rm log}\, p(\bTheta; \vc)= -\lambda \sum_{i,j=1}^{N}w_{ij}\lvert\Theta_{ij}\rvert$ is a \emph{weighted} $\ell_1$-penalty on $\bTheta$ with the weights determined by $\vc$. Thus the proposed optimization problem for learning the model parameters $(\bTheta,\vc)$ is
 \begin{align}
&\underset{\bTheta \succeq 0, \vc}{\text{maximize}\quad}  
\log \operatorname{det} \bTheta - \operatorname{tr}(\mathbf{S} \bTheta) - \lambda \sum\limits_{i,j=1}^{N} w_{ij}\lvert\Theta_{ij}\rvert  
\nonumber \\
&\text{s. to\quad}\, w_{ij} = 1- c_i - c_j + e\, {\rm log}(d_{ij})\nonumber\\
&\quad \quad \quad w_{ij} > 0, \quad i,j=1,2,\ldots,N \nonumber\\
&\quad \quad \quad \sum_{i=1}^N c_i = M, \quad c_i \in [0,1], \quad i=1,2,\ldots, N.
\label{eq:proposed}
\end{align}
The $\ell_1$-penalty in the above optimization problem is relatively smaller if either of the vertices belongs to the core and is the smallest when they both belong to the core as explained by the data. As the value of $e$ is increased (while satisfying the inequality $w_{ij} > 0$), the fraction of edges between spatially distant vertices decreases. Further, to prevent the case where all the weights tend to zero, we constrain the sum of the core scores to a real-valued positive number $M$, while $c_i \in [0,1]$ fixes the scale of the core scores. 

The problem in~\eqref{eq:proposed} is a non-convex optimization problem in the variables $\bTheta$ and $\vc$. In what follows, we propose a solver based on block coordinate ascent to solve~\eqref{eq:proposed}. This decomposes the above non-convex optimization problem into a set of convex subproblems. At each iteration of the algorithm, we update $\bTheta$ by fixing $\vc$ and then update $\vc$ while fixing~$\bTheta$. 

\vspace*{-2mm}
\subsection{Updating the graph \label{subsec:learning theta}} 
For a fixed $\boldsymbol{\vc}$, the problem in~\eqref{eq:proposed} simplifies to the following graphical lasso problem with a weighted $\ell_1$-regularization 
\begin{equation}
\underset{\bTheta \succeq 0}{\text{maximize}\quad} \log \operatorname{det} \bTheta-\operatorname{tr}(\mathbf{S} \bTheta)-\lambda \sum_{i,j=1}^{N}w_{ij}\lvert \Theta_{ij}\rvert
\label{eq:theta step}
\end{equation}
with known weights that depend on $\vc$. This is a convex program that can be solved using existing solvers, e.g., QUIC~\cite{Hsieh2014QUIC}. 

\vspace*{-2mm}
\subsection{Updating the vertex core scores  \label{subsec:learning c}}

For a fixed $\bTheta$, the problem in~\eqref{eq:proposed} simplifies to 
\begin{align}
&\underset{c_1,\cdots,c_N}{\text{maximize}\quad} 
\sum_{i,j=1}^{N}|\Theta_{ij}|(c_i+c_j) \nonumber\\
&\text{s. to \quad}\,
\sum\limits_{i=1}^N c_i = M,\,\, c_i \in [0,1], \,\, \nonumber \\ 
& \quad \quad \quad c_i + c_j < 1 + e {\rm log} (d_{ij}),\, i,j= 1,\cdots,N.
\label{eq:c step}
\end{align}
which is a linear program that can be solved using standard off-the-shelf solvers. We can clearly see from the objective function that the core scores are influenced by the edge weights, which are in turn learnt from data. The subproblem in~\eqref{eq:c step} can also be used to estimate core scores given a graph.

The proposed procedure of block coordinate ascent is initialized with an arbitrary $\vc$ (we use a scaled all-one vector in our experiments) and is repeated till convergence. 


\vspace*{-2mm}
\section{Numerical experiments \label{sec:numerical exp}}

\begin{figure*}[t]
    \centering
        \centering
        \includegraphics[width=1.7\columnwidth]{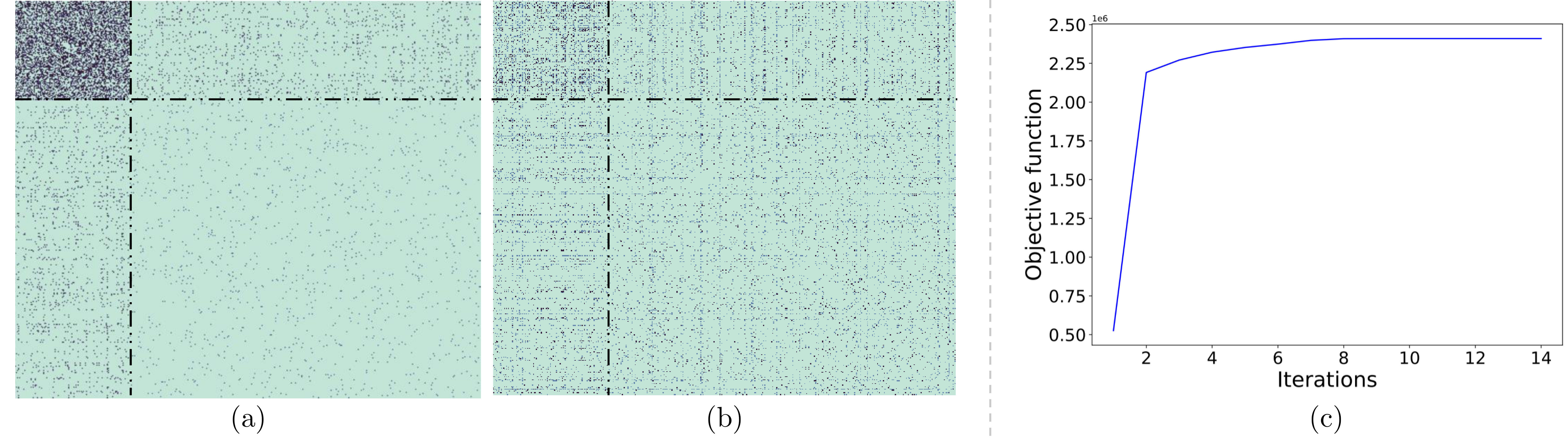}
        \caption{\small (a) Estimated and (b) ground truth networks of the Twitter dataset ordered in the descending order of the estimated core scores. (c) Convergence of the algorithm for \emph{Celegans} dataset.}
        \label{fig:eval}
   \vspace{-4mm}
\end{figure*} 

In this section, we evaluate the graph inference and core score learning capabilities of our framework on real-world datasets from biological, social, and transportation networks. We compare the core scores estimated from existing algorithms, namely, \texttt{MINRES}~\cite{Boyd2010Computing}, \texttt{Rombach}~\cite{Rombach2014Core}, \texttt{RandomWalk}~\cite{Della2013Profiling} and \texttt{k-cores}~\cite{Alvarez2005K}. \texttt{MINRES} learns the core scores $\vc$ such that the adjacency matrix is approximated by $\vc\vc\rT$. \texttt{Rombach} is an extension of the continuous formulation of~\cite{Borgatti2000Models}, which proposes an ideal block model for core-periphery structured networks and estimates core scores by comparing the given adjacency matrix with the ideal block model. \texttt{RandomWalk} estimates the core scores in a network by developing the behavior of a random walker. ~\texttt{k-cores} is a method for partitioning the nodes in a network recursively from the periphery to the more central ones. The input to these existing methods is the ground truth graph. In contrast, the input to our method is just the node attributes.
\vspace{-2mm}
\subsection{Model evaluation} 

We first apply our method on Celegans~\cite{Marcus2006Non}, Cora~\cite{Sen2008Collective}, London underground~\cite{Jia2019CP}, Twitter~\cite{Greene2013Producing}, and WebKB~\cite{Craven1998Learning} datasets. 

Cora is a citation network dataset. In addition to the network of citations, it also contains a binary matrix of size $N\times d$, where $N=2708$ is the number of papers in the dataset and $d=1433$ is the size of the vocabulary. The $(i,j)$th entry of $\mX$ indicates if the $j$th word in the vocabulary is present in the $i$th paper. The Twitter dataset consists of data related to 464 Twitter users, covering athletes and organizations involved in the London 2012 summer olympics. The node attribute matrix $\mX \in \mathbb{R}^{464\times 3097}$ corresponds to the lists of the users, and the ground truth network is formed by the followers' information, with an edge between two users if either of them follow the other. The WebKB~\cite{Craven1998Learning} dataset contains webpages collected from computer science departments of four universities. The node attribute matrix $\mX\in \mathbb{R}^{4518\times d}$ is a binary matrix indicating whether each word in the vocabulary is present or absent in the webpage. The vocabulary size is $d=1703$ words. The spatial distance information for Cora, Twitter, and WebKB datasets is not available.

The network in Celegans dataset is a network of neurons and synapses in a type of worm, called \emph{C. elegans}. The dataset also contains the spatial coordinates of the nodes in the network. In addition to using them as $d_{ij}$, we also consider the coordinates as the data matrix $\mX \in \mathbb{R}^{N\times 2}$, where $N=131$ is the number of nodes in the network. London underground dataset is a network of an underground transportation system. The spatial coordinates of tube stations form the data matrix $\mX \in \mathbb{R}^{N\times 2}$, where $N=315$ is the number of tube stations. For Cora and Twitter datasets, the hyperparameter $e$ is set to $0$ and to 0.09 for Celegans and London underground datasets. We fix $M$ to $N/8$ for all the datasets. Increasing $M$ increases the fraction of nodes with high core scores, whereas $\lambda$ can be tuned according to the required percentage of edges in the network.


Once the core scores are learnt, we order the nodes of both the learnt and the ground truth networks in the decreasing order of core scores. We often observe a perfect core-periphery structure in the estimated network. Although the proposed method is agnostic of the ground truth network, we observe that the ground truth network, when ordered according to the core scores learnt using the node attributes of the network alone, reveals the core-periphery structure. As an example, the adjacency matrices of the learnt and the ground truth networks of Twitter dataset are shown in Fig.~\ref{fig:eval}. The adjacency matrices computed using the proposed method and graphical lasso on a subset of WebKB data related to Texas University are shown in Fig.~\ref{fig:webpage}(a) and Fig.~\ref{fig:webpage}(b), respectively. 


\begin{table*}
\begin{tabularx}{\linewidth}{l l Y Y Y Y Y} \cline{1-7}
    	                  & 	                                                            &  \text{Proposed} & \texttt{MINRES} & \texttt{Rombach} & \texttt{RandomWalk} & \texttt{k-cores} \tabularnewline [0.5ex] \cline{1-7}
 \text{Celegans}          & $\|\bTheta_0-\bTheta_{\rm ideal}\|_F^2$   & $41.940$ & $41.821$ & $39.076$ & $40.877$ & $39.051$ \tabularnewline 
		                  & $\| |\bTheta| -\bTheta_{\rm ideal}\|_F^2$ & $32.642$ & $32.841$ & $32.538$ & $32.707$ & $32.748$ \tabularnewline\cline{1-7}

 \text{Cora}              & $\|\bTheta_0-\bTheta_{\rm ideal}\|_F^2$  & $55.488$ & $55.434$ & $54.690$ & $55.326$ & $54.909$ \tabularnewline 
		                  & $\| |\bTheta| -\bTheta_{\rm ideal}\|_F^2$ & $47.626$ & $55.722$ & $47.884$ & $47.983$ & $47.625$ \tabularnewline\cline{1-7}

\text{London underground} & $\|\bTheta_0-\bTheta_{\rm ideal}\|_F^2$   &   $79.216$ & $79.249$ &  $78.7563$ & $79.338$ & $79.169$ \tabularnewline 
		                  & $\| |\bTheta| -\bTheta_{\rm ideal}\|_F^2$ & $78.818$ & $78.905$ & $78.811$ & $78.858$ & $78.856$ \tabularnewline\cline{1-7}

 \text{Twitter}           & $\|\bTheta_0-\bTheta_{\rm ideal}\|_F^2$  & $134.692$ & $137.142$ & $124.112$ & $131.278$ & $129.221$ \tabularnewline 
		                  & $\| |\bTheta| -\bTheta_{\rm ideal}\|_F^2$ & $110.526$ & $111.837$ & $111.427$ & $111.429$ & $110.526$ \tabularnewline\cline{1-7} \end{tabularx} 
\caption{\small Comparison of the proposed method with the existing core score estimation algorithms.}
\label{table:comparison}
\vspace*{-1mm}
\end{table*}

\begin{figure}
        \centering
        \includegraphics[width=\columnwidth]{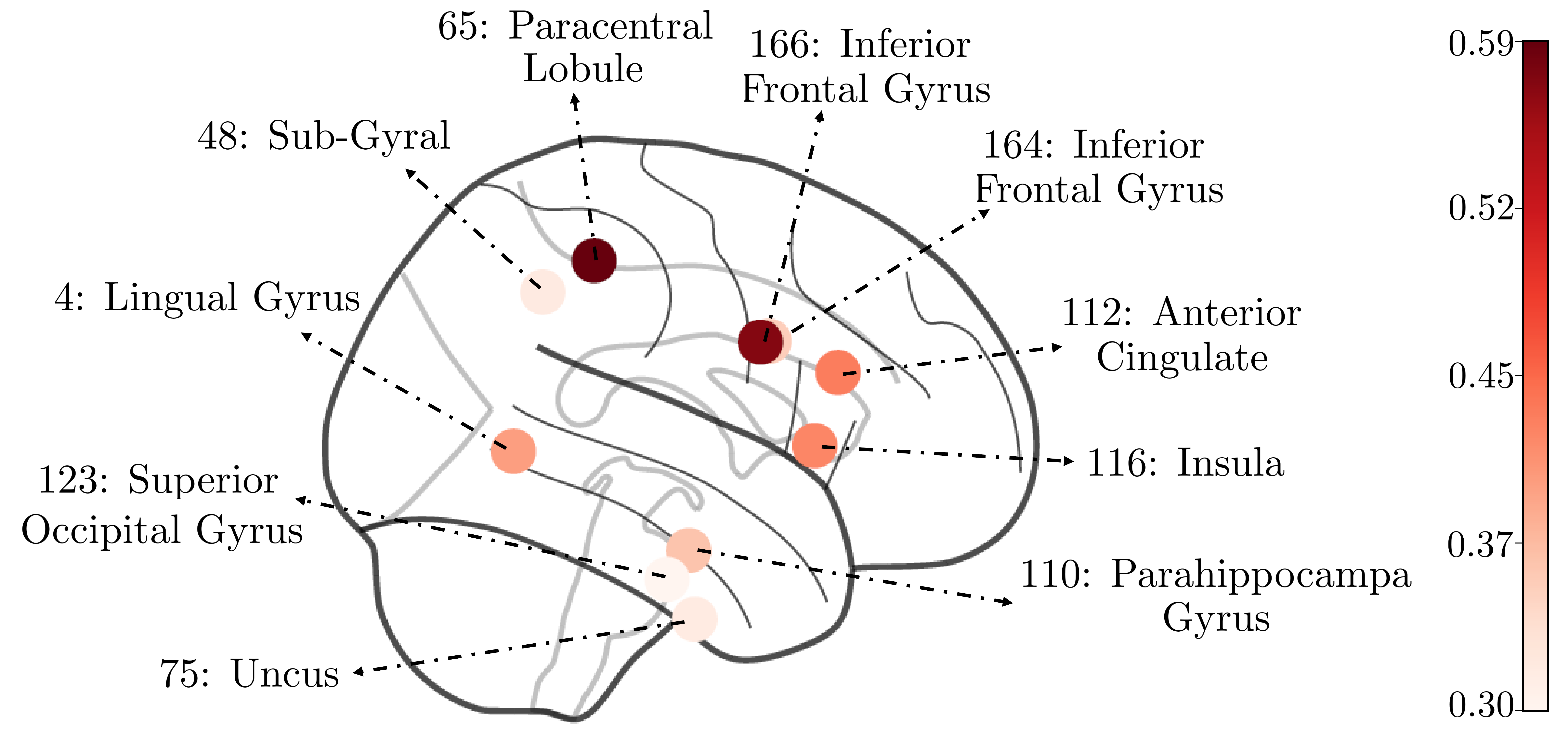}
        \caption{\small Regions with significant differences in the core scores of healthy individuals and subjects with ADHD.}
        \label{fig:brain}
\end{figure}

To compare the proposed method with existing works that estimate core scores given a network, we apply existing core score estimation algorithms on the ground truth networks of the considered datasets, whereas for the proposed method we compute core scores from node attributes. We then order the networks according to the core scores given by the respective algorithms. We compare the ordered networks $\bTheta_0$ with the ideal core-periphery block model~\cite{Borgatti2000Models}, given by 
\begin{equation}
\bTheta_{\rm ideal}= \left[\begin{array}{l|l}
\mathbf{1}_{ tt} & \mathbf{0}_{ t(N-t)}\\
\hline \mathbf{0}_{ (N-t)t} & \mathbf{0}_{ (N-t)(N-t)}
\end{array}\right],
\label{eq:ideal}
\end{equation}
where $\mathbf{1}_{ mn}$ and $\mathbf{0}_{ mn}$ are $m\times n$ dimensional matrices with all ones and all zeros, respectively. We compute $\|\bTheta_0-\bTheta_{\rm ideal}\|_F^2$ for different algorithms. In~\eqref{eq:ideal}, $t$ fixes the proportion of core nodes in the considered network. We fix $t$ to $N/4$ for all the experiments in this section. Comparison of $\|\bTheta_0-\bTheta_{\rm ideal}\|_F^2$ and $\| |\bTheta| -\bTheta_{\rm ideal}\|_F^2$ for different algorithms including the proposed method are shown in Table~\ref{table:comparison}. We observe that the values given by the proposed method are similar to those obtained from the other methods. This indicates that the core-periphery partitioning of the networks by the proposed method is similar to the others, in spite of not knowing the network directly. Furthermore, $\|\bTheta_0-\bTheta_{\rm ideal}\|_F^2 > \| |\bTheta| -\bTheta_{\rm ideal}\|_F^2$  for all the datasets, indicating a more prominent core-periphery structure in the estimated graph than the ground truth, which is also evident from Fig.~\ref{fig:eval}(a). This suggests that the graph estimated using the proposed method by itself can be used to differentiate the core nodes from the peripheral ones.

Finally, Fig.~\ref{fig:eval}(c) shows the convergence plot for the Celegans dataset. We observe that the value of the objective function monotonically increases till convergence. The proposed algorithm converges in less than $10$ iterations for this dataset. 

\vspace*{-2mm}
\subsection{Brain network analysis}

We next apply the proposed method to examine differences between the core and the peripheral regions of healthy individuals and subjects with ADHD.  For this purpose, we use fMRI data from the OHSU brain institute~\cite{Richards2015NITRC}. The dataset $\cbX \in \mathbb{R}^{190\times 74 \times 79}$ consists of fMRI time series for the regions of interest in the cc200 parcellation for a total of  79 individuals, 42 of which correspond to healthy subjects and the others to subjects with ADHD.

We independently compute the core scores of different individuals from their fMRI data. We denote the average of the core score vectors of healthy subjects by $\bar{\vc}_{\rm HC}$ and that of subjects with ADHD by $\bar{\vc}_{\rm ADHD}$. The magnitude of difference between the normalized average core score vectors of the two groups, $|\bar{\vc}_{\rm HC}-\bar{\vc}_{\rm ADHD}|$, serves as a measure of the differences in coreness of different brain regions across the two groups. The highest difference in connectivity (denoting difference in interactions) in the estimated networks is observed in the following regions: \emph{paracentral lobule, inferior frontal gyrus, anterior singulate} and \emph{insula}. Fig.~\ref{fig:brain} shows 10 regions with the largest difference in connectivity, as measured by the 10 largest values of  $|\bar{\vc}_{\rm HC}-\bar{\vc}_{\rm ADHD}|$. The darker nodes in the figure denote the regions with a larger difference in the cores scores of the two groups.
These identified regions, namely, \emph{paracentral lobule, inferior frontal gyrus, anterior cingulate} and \emph{insula} coincide with those reported in~\cite{Dickstein2014The} as the regions with differences in activation for healthy individuals and patients with ADHD. 

\vspace*{-2mm}
\section{Conclusions}
We developed a generative model to relate node attributes to the core scores of vertices through a latent graph structure. Based on the proposed generative model, we presented a joint estimator to simultaneously infer the vertex core scores and a sparse graph whose sparsity pattern is determined by the core scores. The recovered graphs can be readily used to perform core-periphery detection. We presented a block coordinate ascent algorithm to solve the proposed estimation problem. We demonstrated via numerical experiments that the proposed method learns a core-periphery structured graph from only the node attributes while learning core scores on par with methods that use the ground truth network as input. We also applied our method to fMRI data to infer the regions that are the most affected in subjects with ADHD.

\newpage

\bibliographystyle{IEEEtran}
\bibliography{references}
\end{document}